\documentclass[wcp]{jmlr}

\usepackage{longtable}

\usepackage{booktabs}

\title{Unsupervised model compression for multilayer bootstrap networks}

  \author{\Name{Xiao-Lei Zhang} \Email{huoshan6@126.com}\\
  \addr Department of Computer Science and Engineering, The Ohio State University, Columbus, OH 43210, USA.
 }

\begin{document}

\maketitle

\begin{abstract}
Recently, multilayer bootstrap network (MBN) has demonstrated promising performance in unsupervised dimensionality reduction. It can learn compact representations in standard data sets, i.e. MNIST and RCV1. However, as a bootstrap method, the prediction complexity of MBN is high. In this paper, we propose an unsupervised model compression framework for this general problem of unsupervised bootstrap methods. The framework compresses a large unsupervised bootstrap model into a small model by taking the bootstrap model and its application together as a black box and learning a mapping function from the input of the bootstrap model to the output of the application by a supervised learner.
 To specialize the framework, we propose a new technique, named compressive MBN. It takes MBN as the unsupervised bootstrap model and deep neural network (DNN) as the supervised learner. Our initial result on MNIST showed that compressive MBN not only maintains the high prediction accuracy of MBN but also is \textbf{over thousands of times} faster than MBN at the prediction stage. Our result suggests that the new technique integrates the effectiveness of MBN on unsupervised learning and the effectiveness and efficiency of DNN on supervised learning together for the effectiveness and efficiency of compressive MBN on unsupervised learning.
\end{abstract}
\begin{keywords}
Model compression, multilayer bootstrap networks, unsupervised learning.
\end{keywords}

\section{Introduction}

Dimensionality reduction is a core problem of machine learning, where classification and clustering can be regarded as its special cases that reduce high dimensional data to discrete points. In this paper, we focus on unsupervised learning. Traditionally, dimensionality reduction can be categorized to kernel methods, neural networks, probabilistic models, and sparse coding. Kernel methods are too costly on large-scale problems. Although neural networks are scalable to large scale data, they double the computational complexity by a bottleneck structure and take the input as the output of the bottleneck structure at the training stage which is slow, moreover, they learn data distribution globally which is not very effective on learning local structures.

Multilayer bootstrap network (MBN) is a recently proposed bootstrap method (or unsupervised ensemble method). It has multiple nonlinear layers. Each layer is an ensemble of $k$-centers clusterings. The centers of each $k$-centers clustering are only randomly sampled data points (called a bootstrap sample) from the input. MBN is easily implemented and trained, and scales well to large-scale problems as neural networks at the training stage. Moreover, MBN learns a data distribution locally so that it can learn effective representations of data easily. However, MBN contains hundreds of clusterings, which is difficult to be used for prediction.

Motivated by the aforementioned problem and the recent progress of compressing ensemble classifiers to a single small classifier in supervised learning (\cite{bucilua2006model,hinton2015distilling}), in this abstract paper, we propose an unsupervised model compression framework. The framework uses a supervised model to approximate the mapping function from
the input of an unsupervised bootstrap method to the output of the application of the unsupervised bootstrap method. We further specify the framework by taking MBN as the unsupervised bootstrap method and DNN as the supervised model.
The proposed method is named \textit{compressive MBN}. To our best knowledge, this is the first work of model compression for bootstrap methods on unsupervised learning.

\section{Methods}
 Compressive MBN is as follows:
\begin{itemize}
  \item The first step trains MBN on a give training set, and outputs the low dimensional representation of the training data points.
  \item $[$A step driven by applications$]$ The second step applies the low dimensional representation to a given application in unsupervised learning, and outputs the prediction result of the training data.
  \item The third step trains a DNN with the training set as the input and the prediction result as the target. Finally, the DNN model will be used for prediction.
\end{itemize}

The algorithm is a very basic framework. We can easily extend compressive MBN to other techniques by simply using other unsupervised bootstrap techniques to replace MBN in the first step for potentially better performance.
We can also use many other supervised learners to replace DNN in the third step, but to our knowledge, DNN is currently already a good choice.

We may also design a lot of new algorithms by simply specifying the second step for different applications. Some examples are as follows. (i) When compressive MBN is used for visualization, we may omit the second step, and simply take the input and output of the MBN as the input and output of DNN respectively. (ii) When compressive MBN is used for unsupervised prediction, we may run a hard clustering algorithm on the training set, and get the predicted indicator vector of each training data point. For example, if a data point is assigned to the second cluster, then its predicted indicator vector is $[0,1,0,0,\ldots,0]$. We may also get the probabilistic output of the clustering.

\section{Experiments}
We conducted an initial experiment on MNIST. We showed that the technique is very helpful for reducing the high computational cost of MBN on unsupervised prediction problems. The MNIST data was normalized by dividing its entries by 255.

   \begin{figure}[t]
 \centering
       \includegraphics[width=8.3cm]{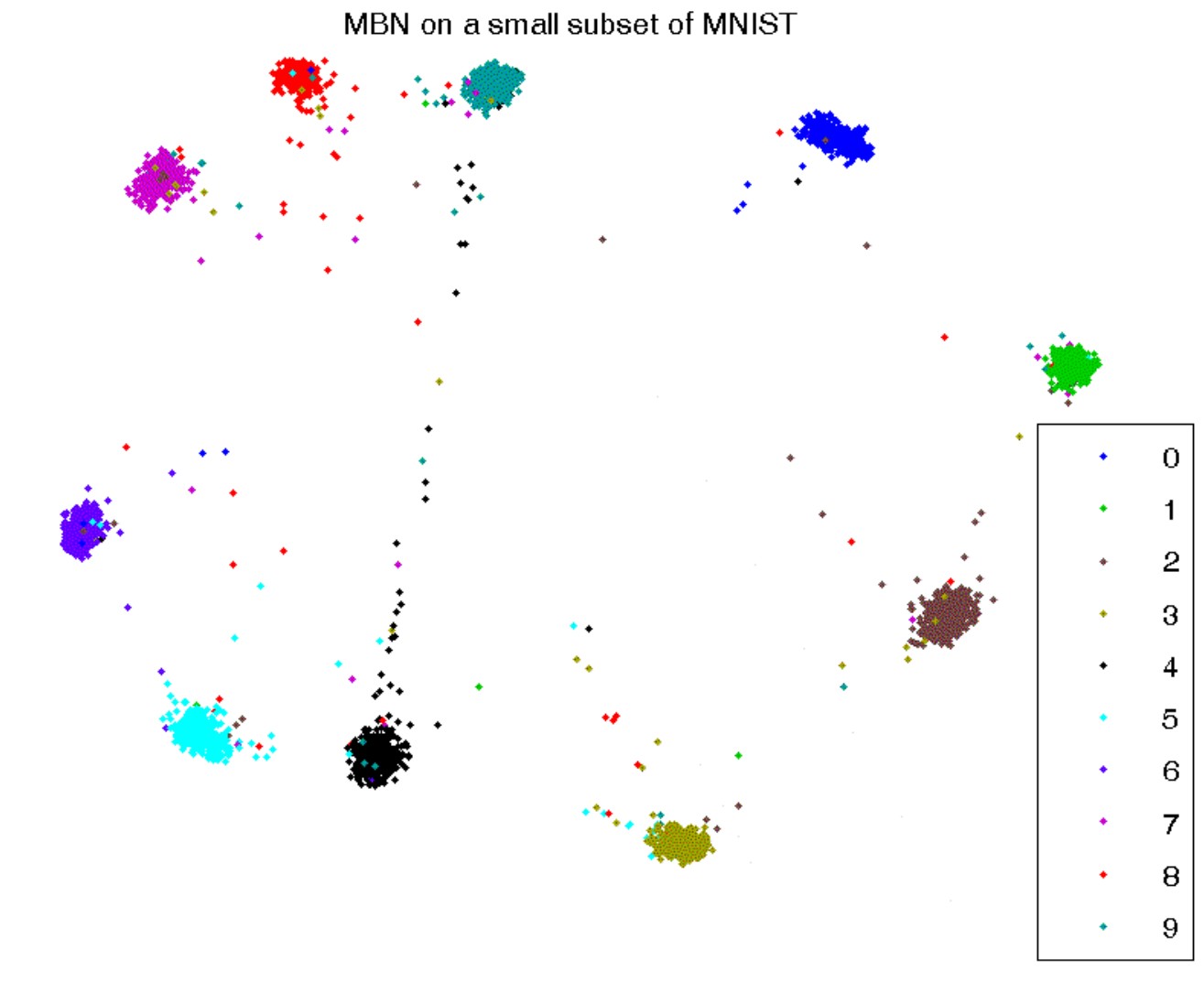}
 \caption{{MBN. Its prediction time on the 5000 images is 2333.24 seconds.}}
 \label{fig:1}
 \end{figure}

  \begin{figure}[t]
 \centering
       \includegraphics[width=8.3cm]{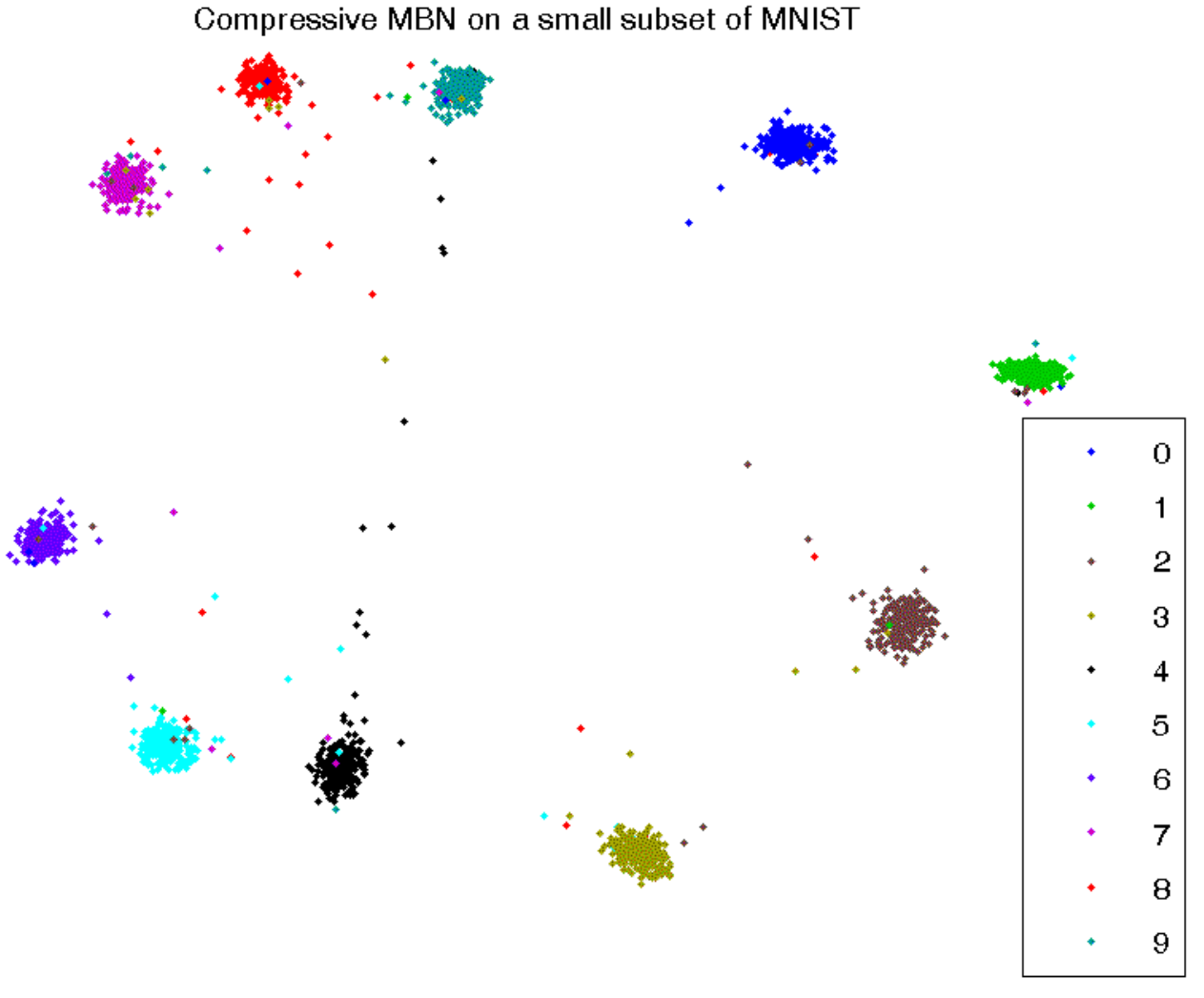}
 \caption{{Compressive MBN. Its prediction time on the 5000 images is 0.58 seconds.}}
 \label{fig:2}
 \end{figure}

\subsection{Experiment on visualizing small subsets of MNIST}\label{subsec:21}
In this subsection, we did not consider the generalization ability of MBN and compressive MBN. Instead, we studied their visualization ability. A data set that contained 5000 unlabeled images randomly selected from the training set of MNIST was used for both training and test.

For the MBN training, we trained MBN similarly as in the second experiment in (\cite{zhang2014nonlinear}). Specifically, the number of clusterings in each layer was set to 400. The parameters $k$ from layer 1 to layer 9 were set to $4000$-$2000$-$1000$-$500$-$250$-$125$-$65$-$30$-$15$ respectively. As we can see, MBN is a very large sparse model: ($4000$-$2000$-$1000$-$500$-$250$-$125$-$65$-$30$-$15$)$\times 400$. The parameter $a$ for random feature selection was set to 0.5. The parameter $r$ for random reconstruction was set to 0.5. After getting the high-dimensional sparse representation from MBN at the top hidden layer, we mapped it to two dimensional space by the expectation-maximization principle component analysis (EM-PCA) (\cite{roweis1998algorithms}).

For the training of compressive MBN, we \textbf{omitted the second step}, and used the input and output of MBN as the input and output of a DNN model respectively. The parameter settings are as follows. We trained a 786-2048-2048-2 DNN. The dropout rate was set to 0.2. The rectified linear unit was used as the hidden unit, and the linear function was used as the output unit. The number of the training epoches was set to 120. The batch size was set to 32. The learning rate was set to 0.001.

 The two dimensional visualizations produced by MBN and compressive MBN were shown in Fig. 1 and Fig. 2 respectively. From the two figures, we found that the visualizations of both MBN and compressive MBN by DNN were equivalently good. When we used the features for clustering, the NMIs of both the methods were around 81\%. Amazingly, \textbf{the prediction time of the compressive MBN on the 5000 images was only 0.58 seconds, which accelerated the prediction time of MBN by around 4000 times!}\footnote{MBN did not enable parallel computing.}

 \subsection{Experiment on the full MNIST}
 We used all 60,000 training images for unsupervised model training and 10,000 test images for test. We discussed the unsupervised generalization ability of the compressive MBN on the test images.

\subsubsection{Compressive MBN without random reconstruction (i.e. parameter $r=0$)}\label{subsubsec:random}
For the MBN training, we trained MBN similarly as in the third experiment in (\cite{zhang2014nonlinear}). Specifically, the number of clusterings in each layer was set to 400. The parameters $k$ from layer 1 to layer 9 were set to $4000$-$2000$-$1000$-$500$-$250$-$125$-$65$-$30$-$15$ respectively. As we can see, MBN is a very large sparse model: ($4000$-$2000$-$1000$-$500$-$250$-$125$-$65$-$30$-$15$)$\times 400$. The parameter $a$ for random feature selection was set to 0.5. The parameter $r$ for random reconstruction was set to 0. After getting the high-dimensional sparse representation from MBN at the top hidden layer, we mapped it to 5 dimensional space by EM-PCA (\cite{roweis1998algorithms}). We further encoded the 5-dimensional representations to 10-dimensional indicator vectors by $k$-means clustering, which was a specialization of the second step of compressive MBN.

  \begin{figure}[t]
 \centering
       \includegraphics[width=8.3cm]{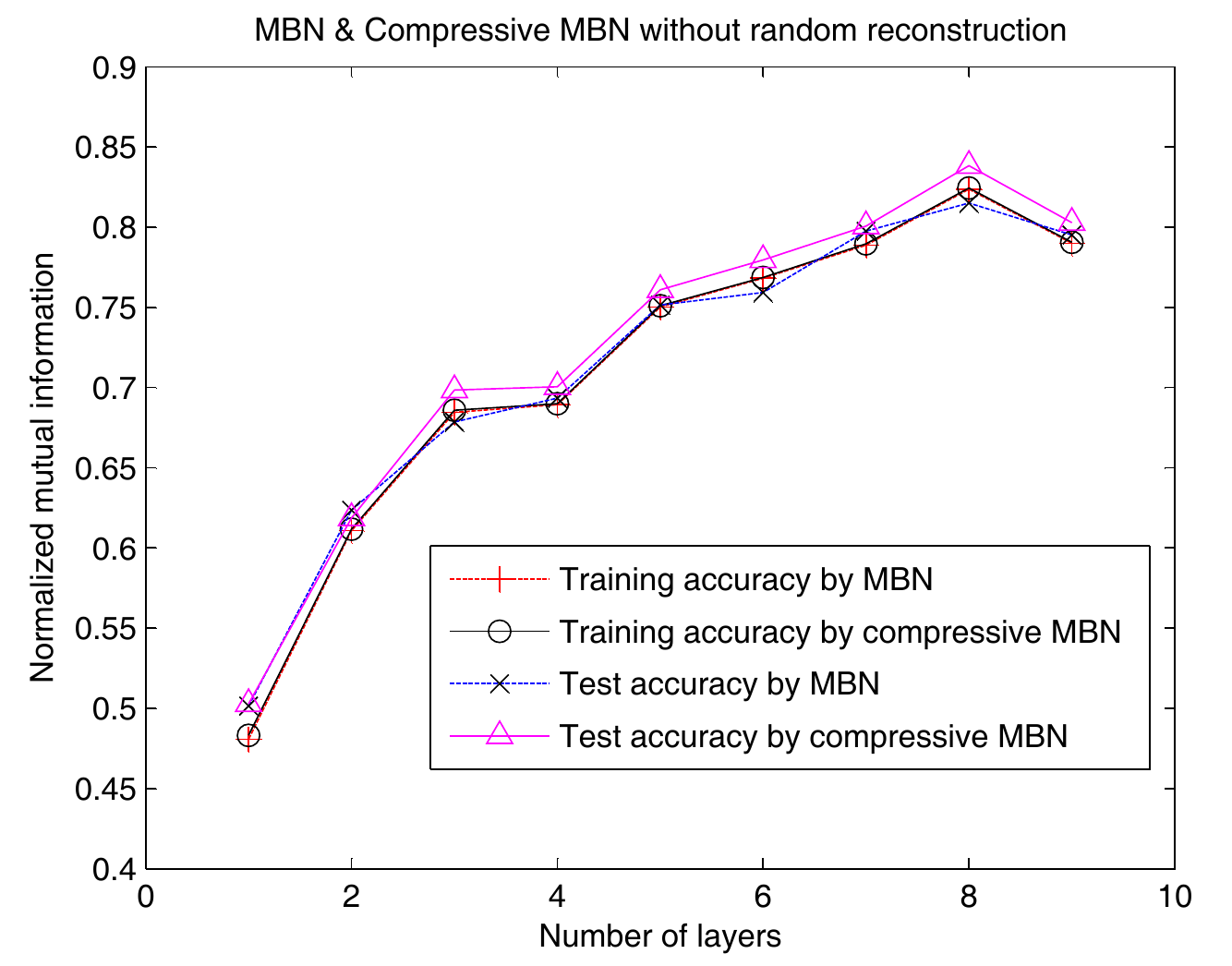}
 \caption{{Comparison of the generalization ability of MBN and compressive MBN on clustering when the random reconstruction of MBN is not used (i.e. $r=0$). The clustering accuracy is evaluated by normalized mutual information. The prediction time of MBN on the 10,000 test images is 4699.69 seconds. The prediction time of compressive MBN on the 10,000 test images is 1.15 seconds.}}
 \label{fig:1}
 \end{figure}

  \begin{figure}[t]
 \centering
       \includegraphics[width=8.3cm]{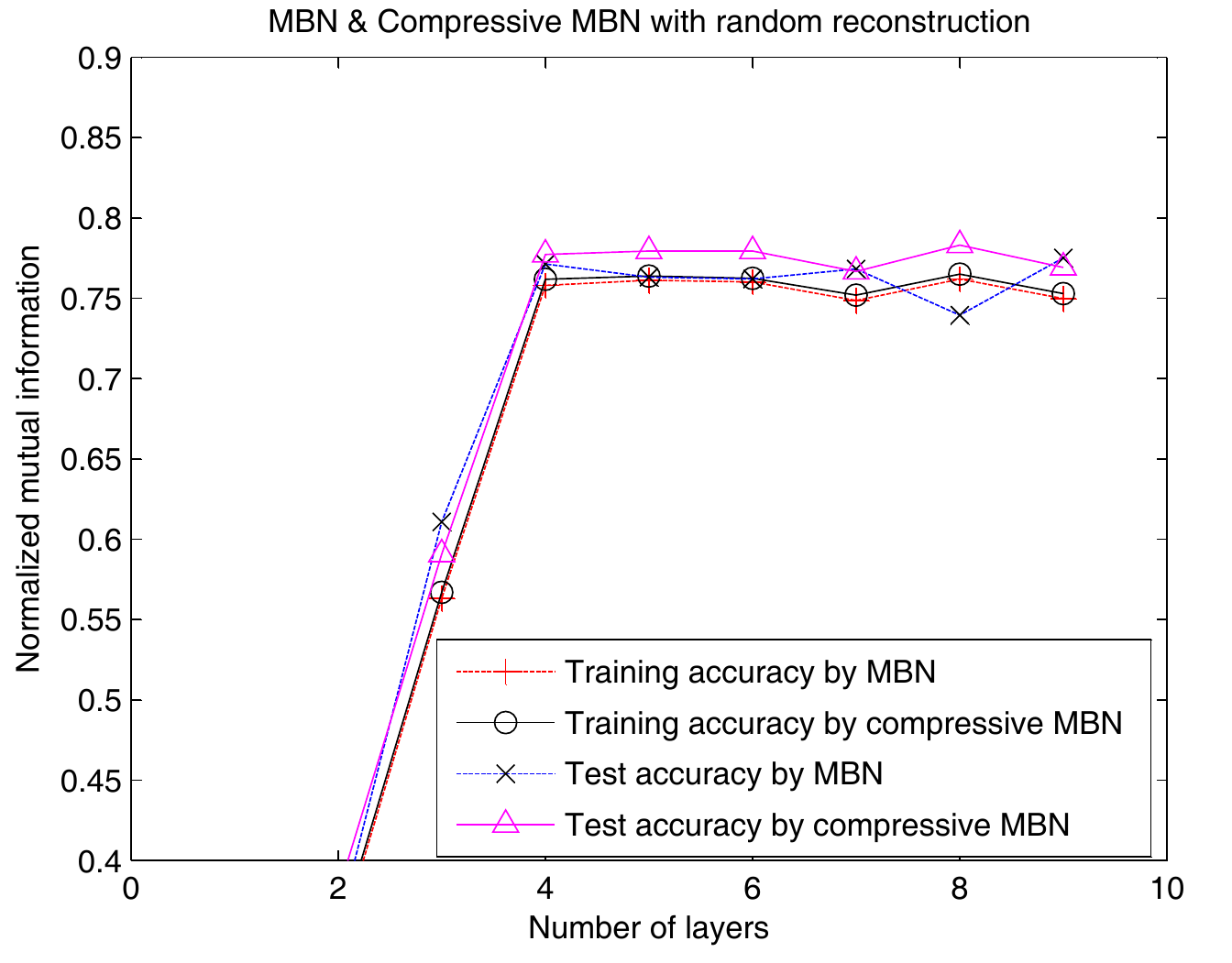}
 \caption{{Comparison of the generalization ability of MBN and compressive MBN on clustering when the random reconstruction of MBN is not used (i.e. $r=0.5$). The prediction time of MBN on the 10,000 test images is 4857.45 seconds. The prediction time of compressive MBN on the 10,000 test images is 1.10 seconds.}}
 \label{fig:2}
 \end{figure}

For the training of compressive MBN, we took the raw feature of the training set as the input of DNN, and took the 10-dimensional predicted indicator vectors as the training target of DNN. The parameter settings of the DNN were as follows. We trained a 786-2048-2048-10 DNN. The dropout rate was set to 0.2. The rectified linear unit was used as the hidden unit, and the sigmoid function was used as the output unit. The number of the training epoches was set to 50. The batch size was set to 128. The learning rate was set to 0.001.

Because $k$-means clustering suffers from local minima, we ran the aforementioned methods 10 times and recorded the average results. The experimental comparison between MBN and compressive MBN was summarized in Fig. 3. From the figure, we observed that the curves of the training accuracy of MBN and compressive MBN were completely coincident; moreover, the curve of prediction of compressive MBN was even \textbf{slightly better} than that of MBN; the highest prediction accuracy of compressive MBN reached 84\% in terms of NMI. The most advanced property of compressive MBN is that it needed only 1.15 seconds to predict 10,000 images, while MBN needed 4699.69 seconds to predict 10,000 images. The prediction time was accelerated by around 4000 times.

\subsubsection{Compressive MBN with random reconstruction (i.e. parameter $r=0.5$)}
It is shown in (\cite{zhang2014nonlinear}) that when the data is small scale (i.e. the training size was similar to the largest parameter $k$), the random reconstruction operation can be quite helpful, however, it is still unclear that whether random reconstruction will be helpful when the data is large scale (i.e. the training size is much larger than the largest parameter $k$), since the largest $k$ in the third experiment of (\cite{zhang2014nonlinear}) was only 1000 and the experimental results of MBN with or without random reconstruction was not very exciting. In this subsection, we enlarged $k$ to 4000 as in Section \ref{subsubsec:random}.

The experimental settings of both MBN and compressive MBN were the same as in Section \ref{subsubsec:random} except that we set $r=0.5$ and mapped the sparse features to 2 dimensional space by EM-PCA. The experimental results were summarized in Fig. 4. From the figure, we observed that all experimental conclusions in Section \ref{subsubsec:random} could also be summarized here, except that when random reconstruction was used, the performance of both MBN and compressive MBN was not as good as that without random reconstruction.

\section{Conclusions}
In this paper, we proposed a general framework for unsupervised model compression. The framework takes MBN  (\cite{zhang2014nonlinear}) as a case of study. The specialized technique, named compressive MBN, uses DNN as an auxiliary model for modeling the mapping function from the input of MBN to the prediction result of a given application that takes the low dimension output of MBN as its input. The new technique aims to solve the problem that
although MBN is simple, effective, robust, and efficient-at-the-training-stage, it is time consuming on prediction.

Our initial experimental result on MNIST showed that compressive MBN not only inherited the generalization ability of MBN (and is even slightly better than MBN), but also accelerated the prediction efficiency of MBN by over thousands of times.

 Compressive MBN concatenates the effectiveness of MBN on unsupervised learning and the effectiveness and efficiency of DNN on supervised learning together for both its effectiveness and its efficiency on unsupervised learning. Moreover, we can easily extend compressive MBN to other unsupervised model compression techniques.

  \section{Acknowledgements}
The author thanks Dr Yuxuan Wang for providing the well-designed DNN toolbox and Prof DeLiang Wang for providing computing resources of the Ohio Supercomputing Center.


\bibliography{zxlrefs}

\end{document}